\documentclass{article} % For LaTeX2e
\usepackage[preprint]{colm2026_conference}

\usepackage{microtype}
\usepackage{graphicx}
\usepackage{subcaption}
\usepackage{booktabs} % for professional tables

\usepackage{pdfcomment}

\usepackage{siunitx}
\usepackage{hyperref}
\usepackage{multirow} % for multi-row table cells
\usepackage{hyperref}
\usepackage{url}
\usepackage{booktabs}
\usepackage{amsmath}
\usepackage{amssymb}
\usepackage{mathtools}
\usepackage{amsthm}
\usepackage{xurl}

\usepackage{lipsum}
\usepackage{float}
\usepackage[most]{tcolorbox}
\usepackage{listings}
\usepackage{xcolor}
\tcbuselibrary{listings, skins, breakable}
\usepackage[T1]{fontenc}  
\usepackage{textcomp}
\usepackage{adjustbox}

\usepackage{array}
\usepackage{wrapfig}

\usepackage[capitalize,noabbrev]{cleveref}
\definecolor{gaingreen}{RGB}{34,139,34}  % 深一点的绿
\definecolor{lossred}{RGB}{178,34,34}    % 深一点的红
\newcommand{\gain}[1]{\textbf{\textcolor{gaingreen}{#1}}}
\newcommand{\loss}[1]{\textbf{\textcolor{lossred}{#1}}}

\theoremstyle{plain}

\theoremstyle{definition}

\theoremstyle{remark}

\usepackage{lineno}

\definecolor{darkblue}{rgb}{0, 0, 0.5}
\hypersetup{colorlinks=true, citecolor=darkblue, linkcolor=darkblue, urlcolor=darkblue}

\title{When Users Change Their Mind: Evaluating Interruptible Agents in Long-Horizon Web Navigation}

\author{Henry Peng Zou\textsuperscript{1*},
    Chunyu Miao\textsuperscript{1*}, 
    Wei-Chieh Huang\textsuperscript{1*}, 
    Yankai Chen\textsuperscript{2,3}, 
    Yue Zhou\textsuperscript{1}, \\
    \bf Hanrong Zhang\textsuperscript{1},
    \bf Yaozu Wu\textsuperscript{1},
    \bf Liancheng Fang\textsuperscript{1},
    \bf Zhengyao Gu\textsuperscript{1},
    \bf Zhen Zhang\textsuperscript{4} \\
    \bf Kening Zheng\textsuperscript{1},
    \bf Fangxin Wang\textsuperscript{1},
    \bf Yi Nian\textsuperscript{5},
    \bf Shanghao Li\textsuperscript{1},
    \bf Wenzhe Fan\textsuperscript{1} \\
    \bf Langzhou He\textsuperscript{1},
    \bf Weizhi Zhang\textsuperscript{1},
    \bf Steve Xue Liu\textsuperscript{2,3},
    \bf Philip S. Yu\textsuperscript{1} \\\\
    \textsuperscript{1}University of Illinois Chicago, 
    \textsuperscript{2}McGill University,
    \textsuperscript{3}MBZUAI, \\
    \textsuperscript{4}University of California Santa Barbara, 
    \textsuperscript{5}University of Southern California \\
}

\begin{document}

\ifcolmsubmission
\linenumbers
\fi

\maketitle
\begin{abstract}
As LLM agents transition from short, static problem solving to executing complex, long-horizon tasks in dynamic environments, the ability to handle user interruptions, such as adding requirement or revising goals, during mid-task execution is becoming a core requirement for realistic deployment.
However, existing benchmarks largely assume uninterrupted agent behavior or study interruptions only in short, unconstrained language tasks. In this paper, we present the first systematic study of interruptible agents in \textit{long-horizon}, \textit{environmentally grounded} web navigation tasks, where actions induce persistent state changes. 
We formalize three realistic interruption types, including addition, revision, and retraction, and introduce \textbf{InterruptBench}, a benchmark derived from WebArena-Lite that synthesizes high-quality interruption scenarios under strict semantic constraints.  
Using a unified interruption simulation framework, we evaluate six strong LLM backbones across single- and multi-turn interruption settings, analyzing both their effectiveness in adapting to updated intents and their efficiency in recovering from mid-task changes.
Our results show that handling user interruptions effectively and efficiently during long-horizon agentic tasks remains challenging for powerful large-scale LLMs. 
Code and dataset are available at \href{https://github.com/HenryPengZou/InterruptBench}{InterruptBench Repository}.

% \href{https://anonymous.4open.science/r/InterruptBench-7BD9/README.md}{Anonymous Github Repository}.

\end{abstract}

\section{Introduction}

The evolution of LLM agents from solving static, instantaneous problems to executing complex, multi-hour tasks in dynamic environments has inaugurated a new era of \textit{realistic} agentic research~\citep{durante2024agent, wu2025multi,erdogan2025plan,huang2026rethinking}. Central to this realism are two critical dimensions: agents must be able to operate over a \textbf{long horizon}, handling hundreds of environmental interactions and issuing large amounts of actions, and must be \textbf{interruptible} to handle dynamic user intent, as users naturally want to examine partial results, correct spotted errors, or revise goals mid-execution over such a long horizon.

Despite the need for interruptible agents, existing research remains remarkably sparse. While recent benchmarks have advanced long-horizon capabilities~\citep{liu2023agentbench,chen2025xbench}, they assume uninterrupted execution. The few works that do consider interruption focus on simple language tasks with severe limitations: they examine only short-horizon scenarios, lack grounded environmental constraints~\citep{wu2025large}, and restrict analysis to single-turn interruptions~\citep{hua2024interactive}. Consequently, a critical gap remains in understanding \textbf{how agents can effectively and efficiently handle dynamic user interruptions in realistic, long-horizon settings} where environmental states are constrained, and actions have lasting consequences.

To address this gap, we present the first systematic study of interruptible agents in long-horizon, environmentally constrained settings. In this work, we focus on web navigation tasks, where agents must execute sequences of grounded actions (clicks, typing, navigation) that produce persistent state changes in dynamic web environments. We identify and formalize three realistic interruption scenarios that capture the spectrum of dynamic user intent: \textit{addition} (adding new requirements to the initial query), \textit{revision} (modifying an incorrect query to fix errors), and \textit{retraction} (removing previously specified details).

To enable rigorous evaluation, we build upon high-quality, human-verified tasks from WebArena-Lite \citep{zhou2024webarena, koh2024visualwebarena, wei2025webagent} and develop a systematic synthesis pipeline to generate realistic interruption scenarios. For each original task, we synthesize an initial user query and subsequent user interruption messages, enforcing strict constraints to enhance data quality: (1) The initial user query combined with the user interruption messages should have the same meaning as the original intent and result in the same ground truth answers; (2) Each interruption message should be essential, lacking any one interruption message should result in different ground truth answers than the original intent; (3) Each interruption message should be expressed in a natural and realistic user/customer tone. Our resulting benchmark, \textbf{InterruptBench}, spans three diverse interruption scenarios and five web domains, with each task requiring agents to handle user interruption messages during long-horizon mid-task execution.

We conduct extensive experiments across six powerful LLM backbones in WebAgent-R1 \citep{wei2025webagent} agent scaffolding, including both open-source and large-scale closed-source models, evaluating their ability to handle interruptions across our three interruption scenarios. 
Our analysis examines not only how effectively agents adapt to updated intents following a user interruption but also how efficiently they do so. Beyond the single-turn interruption scenario, we extend our investigation to a multi-turn interruption scenario in which users issue multiple sequential interruptions during task execution, reflecting realistic interactive patterns over long horizons. 
We find that handling user interruptions effectively and efficiently during long-horizon agentic tasks remains challenging for powerful large-scale LLMs.

Our contributions are summarized as follows:

1. We introduce \textbf{InterruptBench}, the first benchmark for studying how agents handle user interruptions in long-horizon, environmentally constrained settings, grounded in realistic web navigation tasks with persistent state changes.

2. We develop a trajectory-grounded simulation and evaluation framework that injects interruptions at dynamically determined execution points and enables systematic measurement of both \emph{effectiveness} and \emph{efficiency} in post-interruption adaptation.

3. We conduct extensive empirical analysis across six strong LLM backbones under diverse interruption scenarios (addition, revision, retraction) and multi-turn interactions, revealing intriguing insights of interruptible agents.

\begin{figure*}[!t]
    \centering
    \hspace*{2mm}
    \includegraphics[width=\textwidth]{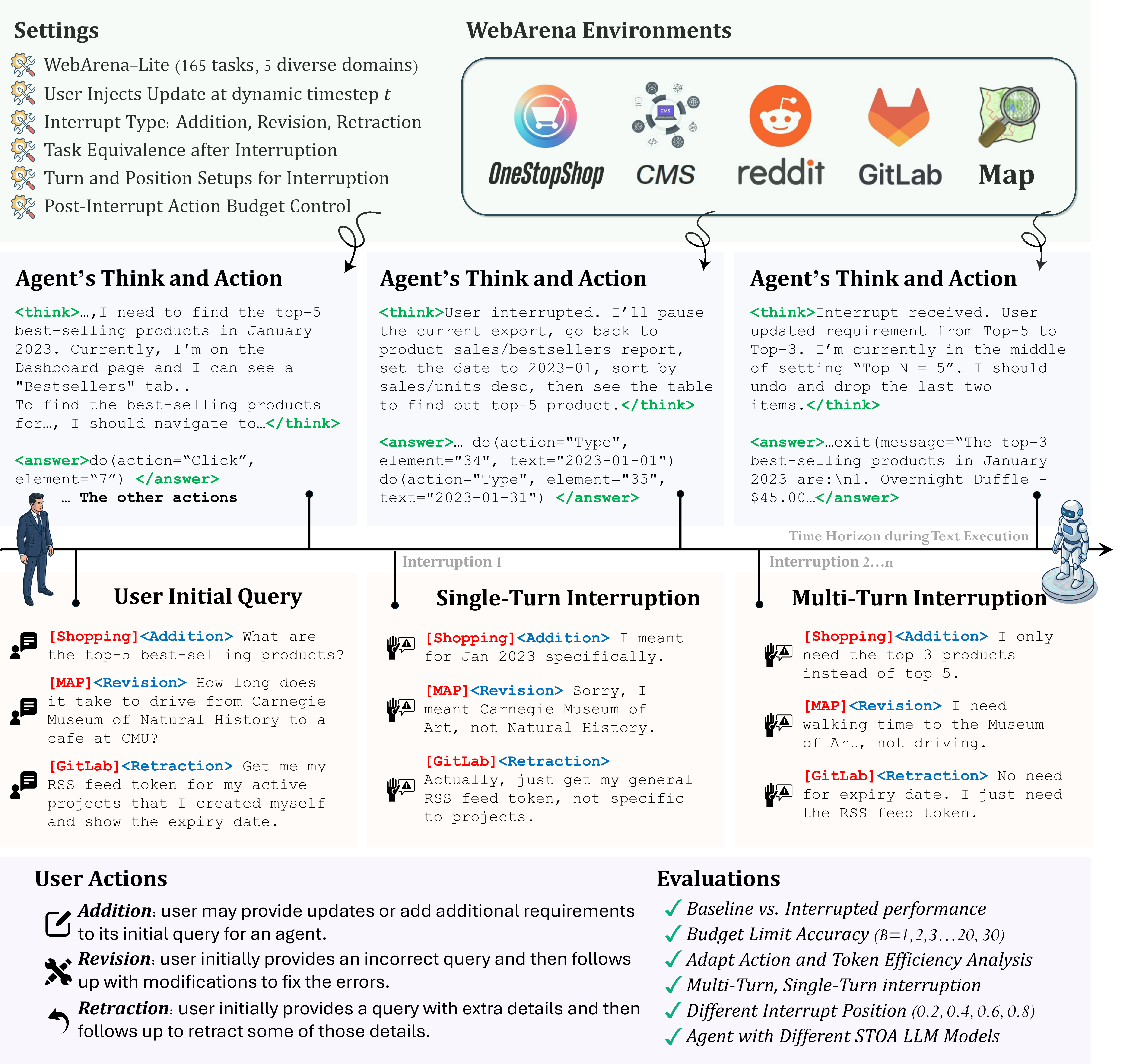} % example-image-duck
    \caption{\textbf{InterruptBench setup and evaluation}. We evaluate an agent operating in the WebArena environment under user-driven interruptions at dynamic timesteps, where the user may update, revise, or retract parts of the original request.}
    \label{fig:main_figure}
\end{figure*}

\section{InterruptBench: Benchmark Construction and Simulation}
% \lipsum
% \section{InterruptBench: Benchmark Construction and Simulation}
% TODO:
% use InterruptBench as benchmark name and refactor corresponding content
% add corresponding dataset statistics?

This section introduces the construction of our interruption benchmark and simulation. We first describe the underlying task suite and environment, then define three realistic mid-task interruption scenarios (Addition, Revision, and Retraction), and finally detail the procedure for simulating these interruptions during agent execution (illustrated in Figure \ref{fig:main_figure}).

% Adaptation to updated intent under increasing post-update action budget.

% Budget-limited successful rate comparison: Success rate as a function of post-interruption action budget \(k\) for three update types (Addition/Revision/Retraction) with interruption position 0.6.
    
\subsection{Task and Environment}

% [Motivation to Choose Webareana/WebAgent-Lite Dataset:] real world agentic long-horizon web tasks with environment constraint (web actions \& states); 
We build our benchmark by augmenting tasks from WebArena-Lite \citep{zhou2024webarena,koh2024visualwebarena, wei2025webagent}. WebArena-Lite provides realistic long-horizon web tasks with reliable constraints on environmental state and programmatic success checking. It covers five diverse domains and environments, including social forum (Reddit), collaborative coding (GitLab), e-commerce content management system (CMS), open street map (Map), and online shopping (OneStopShop), and provides a curated evaluation set of 165 human-verified tasks. Each task in WebArena-Lite requires the agent to interact with the environment via a sequence of observations and actions and can involve up to 30 LLM calls by default. More details about WebArena-Lite are provided in Appendix \ref{appendix:environment_statistics}.
% \cyk{I still suggest to give a name for our bench and then replace the name with our modification introduced here.}. 
% [refer to WebAgent-R1 Appendix B and WebArena paper]. 

\subsection{Interruption Scenarios and Data Synthesis}
\label{sec:interruption_scenario}

In this work, we consider three realistic and most common interruption scenarios:

\textbf{Addition:} An interruption scenario where a user initially provides a basic or incomplete query and then follows up with additional details, new information, or supplementary requirements to expand upon the original request.

\textbf{Revision:} A modification interruption scenario where a user initially provides an incorrect or mistaken query and then follows up with corrections/modifications to fix the errors.

\textbf{Retraction:} A retraction interruption scenario where a user initially provides a query with extra details/constraints and then follows up to remove or retract some of those details.

In addition, we consider a baseline \textbf{No-Interruption} scenario, where the agent receives only the initial user query, without any interruptions or additional follow-up during task execution.

\textbf{Data Synthesis.} For each original task in WebArena-Lite, we prompt a strong LLM to synthesize an initial user query and the additional user interruption messages to be injected during agent task execution. Notably, we enforce the following strict \textit{constraints} to enhance data quality: 

\textbf{(1) Ground Truth Consistency:} The initial user query, combined with the user interruption messages, should have the same meaning as the original query in WebArena-Lite and lead to the same ground truth answer, enabling reliable evaluation.

\textbf{(2) Necessity:} Each interruption message should be essential; lacking any interruption message should result in a different ground truth answer from the original intent.

\textbf{(3) Authenticity:} Each interruption message should be expressed in a natural and realistic user/customer tone. 

The complete prompts for synthesizing each type of interruption data are provided in Appendix \ref{appendix:prompt}. All generated data are manually reviewed by the authors to ensure that each interruption message is necessary for correctly solving the problems.

% \subsection{Trajectory-Grounded Interruption Simulation}
% We simulate interruptible user--agent interactions by injecting mid-task user updates into trajectory-grounded executions.

% \textbf{(1) Generate initial trajectory.} We first execute the initial user query to obtain a full action trajectory and record its length.

% \textbf{(2) Replay with progress-aware interruption.} We then replay the trajectory and inject interruptions based on relative progress rather than fixed steps. Specifically, each interruption is triggered at a predefined proportion of the initial trajectory, ensuring consistent semantic timing across tasks of varying lengths.

% \textbf{(3) Inject interruption and continue execution.} We insert the user update and continue generating the post-interruption trajectory.

% This trajectory-grounded design enables controlled and comparable evaluation of interruption handling across tasks with varying lengths and complexities. We consider both \textit{single-turn} settings with a single interruption and \textit{multi-turn} settings where users iteratively refine their intent through multiple interruptions, reflecting realistic long-horizon interaction patterns.

\subsection{Trajectory-Grounded Interruption Simulation}

We simulate user–agent interactions in which the agent first executes the initial user query and is then interrupted mid-task by additional user messages. Such simulation consists of three steps:

\textbf{(1) Generate initial full trajectory.} 
% We first generate the initial full trajectory using the initial user query and record the total number of actions in this trajectory. 
We first execute the initial user query to obtain a complete trajectory and record its total number of actions.

% \textbf{(2) Replay until interruption step.} We then replay the initial trajectory sequentially and use a dynamic triggering mechanism based on relative trajectory progress rather than a fixed global step. For each task, the interruption is injected at a predefined relative position along that task's initial trajectory, so the exact interruption step is automatically determined by trajectory length and therefore varies across tasks. 
\textbf{(2)  Replay with progress-aware interruption.} We then replay the trajectory sequentially and inject the interruption based on \emph{relative progress} rather than a fixed global step. Specifically, for each task, the interruption is introduced at a predefined proportion of the initial trajectory, so the exact interruption step is automatically determined by the initial trajectory length and varies across tasks.

\textbf{(3) Insert interruption and continue execution.} 
% Finally, we inject the additional user message and continue generating the post-interruption trajectory.
Finally, we insert the additional user message at the selected interruption step and continue generating the post-interruption trajectory conditioned on the updated user intent.

This trajectory-grounded design enables controlled and comparable evaluation of interruption handling across tasks with varying lengths and complexities. 

We consider both \textit{single-turn interruption} settings, in which only one interruption is introduced, and \textit{multi-turn interruption} settings, in which users may interrupt the agent multiple times to provide iterative feedback.

\section{Experimental Setup}

\subsection{Models}
\label{sec:exp:models}
We evaluate six LLM backbones in the WebAgent-R1~\citep{wei2025webagent} agent scaffolding:
Claude-Haiku-4.5~\citep{claude45_hiku}, Claude-Sonnet-4.5~\citep{claude45}, Claude-Opus-4.5~\citep{claude45_opus}, Qwen3-Coder-480B-A35B~\citep{yang2025qwen3} (abbreviated as Qwen3-480B-A35B),
DeepSeek-V3.1~\citep{liu2024deepseek}, and Mistral-Large-3~\citep{jiang2023clip}.
Unless noted, we keep the prompting format, tool/API surface, termination
conditions, and maximum action budget identical across models. We leverage Claude-Opus-4.5 for data synthesis by default. 
% TODO: perhaps add default values

\subsection{Interruption Configurations}
\label{sec:exp:interruptions}
% \paragraph{Interruption scenarios.}
% We study Addition, Revision, and Retraction scenarios defined in Section \ref{sec:interruption_scenario}.

\paragraph{Interruption scenarios and counts.} We study the Addition, Revision, and Retraction interruption scenarios defined in Section \ref{sec:interruption_scenario}. 
% TODO: random
We evaluate both single-interruption episodes and multi-interruption episodes. For multi-interruption runs, we consider \emph{random} sequences, where the interruption type can vary across turns. For example, an Addition followed by a Revision and then a Retraction.
Throughout, interruptions are \emph{informational}: they do not reset the agent or the environment, and do not invalidate progress already made. Instead, they provide additional user-supplied necessary information that the agent can leverage to finish the task more effectively.

\paragraph{Timing strategy.}
We inject interruptions at a relative position (60\% by default) of the baseline trajectory length obtained from the corresponding no-interruption run.
To compare post-interruption adaptation across tasks with different pre-interruption
lengths, we align trajectories using the post-interruption action index $k$,
where $k{=}1$ is the first action after receiving an update. Evaluation under different interruption positions is provided in Appendix~\ref{appendix:eval_position}.

\subsection{Metrics}
\label{sec:exp:metrics}
We evaluate agents along three complementary dimensions: task success, efficiency,
and performance under multiple interruption updates.

% TODO: need to refine according to WebAgent-R1
\paragraph{Task success and post-interruption success curves.}
We define success with respect to the \emph{final} post-update intent. We calculate the Success Rate (SR), which indicates whether a task is successfully completed according to three rule-based evaluation metrics provided in WebAgent-R1 \citep{wei2025webagent}: String Match, URL Match and Program Execution. 
% (More details on these rule-based metrics are provided in Appendix \ref{appendix:environment_statistics}). 
To quantify how agents incorporate updated information over time, we report the post-interruption success curve, denoted as $\mathrm{SR}(k)$ (also referred to as the budget-limited success rate), defined as
\begin{equation}
\mathrm{SR}(k) \,=\, \frac{1}{N}\sum_{i=1}^{N} \mathbb{I}\big[\tau_i \le k\big],
\end{equation}
\label{eqn:SR}
where $\tau_i$ denotes the number of actions taken after the interruption until success is achieved in episode $i$, and $k$ is the post-interruption action index.

% For multi-interruption episodes, $k$ is measured from the \emph{latest} interruption.

\paragraph{Efficiency.}
We measure output tokens and action count; output tokens serve as a proxy for the generation cost of producing actions and incorporating updated user intents.
To isolate the effect of interruptions, we compare each interrupted run against a matched no-interruption baseline on the same underlying task.
We additionally stratify episodes by paired outcomes to support fine-grained analysis.

\paragraph{Performance under multiple interruption updates.}
We report success as a function of the number of interruptions and study how $\mathrm{SR}(k)$ and representative failure patterns evolve as interruptions accumulate.

\section{Experimental Results and Analysis}

\begin{figure*}[t]
    \centering
    \includegraphics[width=\textwidth]{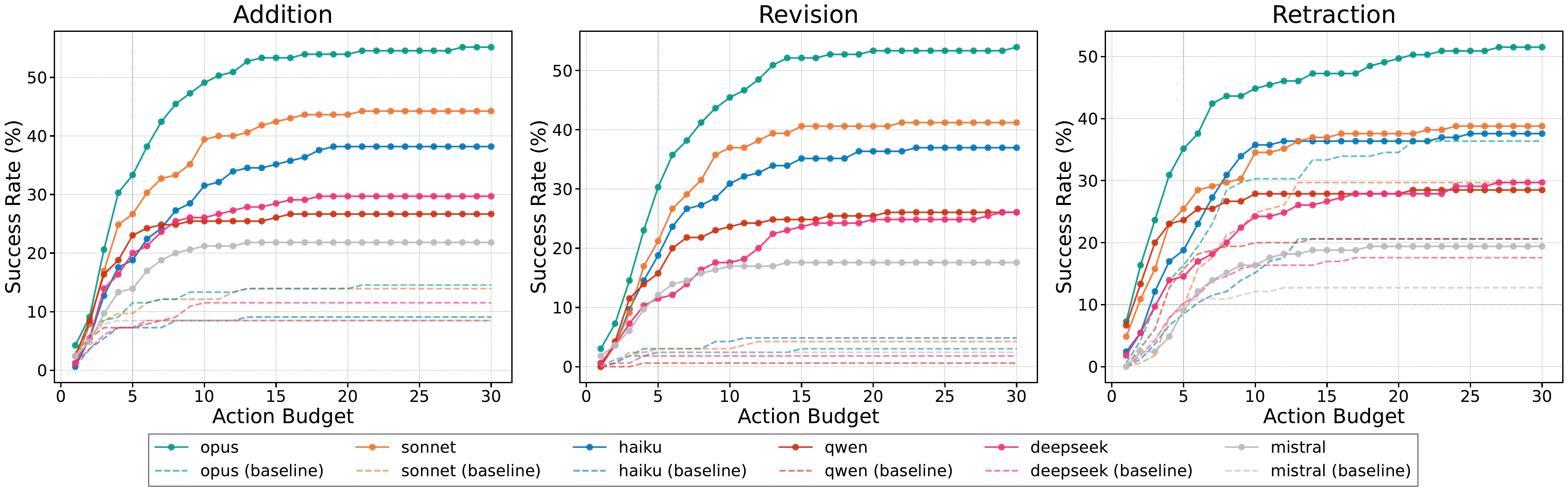}
    % \caption{Budget-limited successful rate comparison: Success rate as a function of post-interruption action budget \(k\) for three
    % interruption scenarios.
    % Solid lines correspond to runs that receive the mid-task interruption update; dashed
    % lines are matched no-interruption runs that do \emph{not} receive the update
    % but are evaluated on the final intent, providing a lower-bound reference for
    % \emph{no-update} behavior under the same budget.}
    \caption{\textbf{Budget-limited success rate as a function of post-interruption action budget \(k\) across three interruption scenarios.} Solid lines denote runs that receive mid-task interruption updates, while dashed lines represent matched no-interruption runs that do \emph{not} receive updates but are evaluated against the final intent, serving as a lower-bound reference for \emph{no-update} behavior under the same budget.}
    \label{fig:recovery_curves}
\end{figure*}

\label{sec:results}
We structure our empirical analysis around three questions:
\textbf{(Q1)} How do agents adapt to updated intents after an interruption?
\textbf{(Q2)} How efficient is their adaptation?
\textbf{(Q3)} How does performance change under multiple interruptions?

\subsection{Adaptation Dynamics After Interruption}
\label{sec:results:q1}
We analyze adaptation by aligning trajectories at the interruption point and computing the post-interruption success curve $\mathrm{SR}(k)$: the fraction of
episodes that achieve the \emph{final} intent within $k$ actions after the interruption.
This produces a family of curves $\mathrm{SR}_{m,t}(k)$ for model $m$ and interruption type $t$.
Figure~\ref{fig:recovery_curves} visualizes $\mathrm{SR}(k)$, where solid lines denote interrupted runs and dashed lines denote matched no-interruption \emph{lower-bound} baselines that do not receive the interruption update but are still evaluated against the final intent.

\paragraph{Observed patterns.}
Figure~\ref{fig:recovery_curves} characterizes adaptation in the \emph{budget-limited} setting, which operationalizes human-in-the-loop corrections under a constrained remaining action budget.
Across all interruption types, success rate increases rapidly with budget up to approximately $k \approx 10$ and then plateaus, indicating that most adaptation occurs within the first few post-update actions.
Model differences are apparent both in the rate of early improvement and in the eventual ceiling.
For the Addition scenario, Opus reaches approximately 50\% by $k = 10$ and 55\% by $k = 30$, while Sonnet reaches roughly 40\% and 44\%, and Haiku reaches approximately 33\% and 38\%.
Open-weight backbones plateau at lower levels; for instance, Qwen reaches approximately 26\%, DeepSeek 30\%, and Mistral 23\% at $k = 30$.

The dashed lower-bound baselines clarify how much of the observed performance depends on incorporating the update.
For Revision, baselines remain near zero across all budgets for most models, indicating that success is largely contingent on correctly integrating the modified information; accordingly, the interrupted curves show substantial absolute gains (e.g., Opus exceeds 50\% by $k \approx 15$).
For Addition, baselines are higher but still well below the interrupted curves, suggesting that additions are partially compatible with the pre-update plan but still require explicit constraint integration.
In contrast, Retraction exhibits notably stronger baselines, resulting in a smaller absolute gap between interrupted and uninterrupted curves; here, the primary benefit of the update is earlier task completion under tight budgets rather than a substantially higher overall ceiling.

% \noindent\textbf{Adaptation is early, scenario-dependent, and model-consistent.}
% Figure~\ref{fig:recovery_curves} presents the post-interruption success curves $\mathrm{SR}(k)$ across all three interruption scenarios.
% Across models and scenarios, success rate rises steeply within the first $k \approx 10$ actions and largely saturates thereafter, indicating that adaptation either occurs quickly or not at all.
% The gap between interrupted runs (solid) and no-interruption baselines (dashed) varies by scenario: in Revision, baselines remain near zero, meaning success depends almost entirely on integrating the corrected information; in Addition, baselines are moderate, reflecting partial compatibility with the original plan; in Retraction, baselines are the strongest, with the update primarily accelerating completion rather than raising the success ceiling.
% Model rankings are consistent across scenarios, with the Claude family consistently outperforming open-weight alternatives and Opus achieving the highest plateau in all three settings.

\begin{table*}[t]
    \centering
    \small
    \setlength{\tabcolsep}{4pt}
    \begin{adjustbox}{max width=\textwidth}
    \begin{tabular}{ll
        S[table-format=-5.2]
        S[table-format=-5.2]
        S[table-format=-5.2]
        S[table-format=-5.2]
        S[table-format=-5.2]
        S[table-format=-5.2]
        S[table-format=-5.2]
    }
        \toprule
        \multicolumn{9}{c}{Scenario: Addition} \\
        \midrule
        Model & Metric & {S/F} & {F/S} & {S/S} & {F/F} & {No-int} & {With-int} & {Avg.} \\
        \midrule
        \multirow{3}{*}{Claude-Haiku-4.5} & \# cases  & 3 & 52 & 11 & 99 & 165 & 165 & \multicolumn{1}{c}{--} \\
         & \# action & 6.6667 & -2.3077 & -1.7273 & 1.1313 & \loss{\tablenum{14.7758}} & \loss{\tablenum{14.7333}} & -0.0424 \\
         & \# token  & 1808.3333 & 330.5962 & -187.6364 & 2624.2559 & 3896.0101 & 5595.1212 & \loss{\tablenum{1699.1111}} \\
        \midrule
        \multirow{3}{*}{Claude-Sonnet-4.5} & \# cases  & 4 & \tablenum{55} & 18 & 88 & 165 & 165 & \multicolumn{1}{c}{--} \\
         & \# action & 21.7500 & -1.8909 & -3.8889 & 2.5341 & 13.2364 & 14.0606 & 0.8242 \\
         & \# token  & 8584.3333 & -123.7697 & -896.5370 & 1128.4205 & 3086.1212 & 3756.9899 & 670.8687 \\
        \midrule
        \multirow{3}{*}{Claude-Opus-4.5} & \# cases  & \gain{\tablenum{1}} & \gain{\tablenum{70}} & \gain{\tablenum{21}} & \gain{\tablenum{73}} & 165 & 165 & \multicolumn{1}{c}{--} \\
         & \# action & 0.0000 & -2.2143 & -1.8571 & 0.4110 & \gain{\tablenum{11.0909}} & \gain{\tablenum{10.0970}} & \gain{\tablenum{-0.9939}} \\
         & \# token  & 312.0000 & -92.7000 & 159.2857 & 350.5297 & 1635.0828 & 1773.0020 & 137.9192 \\
        \midrule
        \multirow{3}{*}{Qwen3-235B-A22B} & \# cases  & 2 & 32 & 12 & 119 & 165 & 165 & \multicolumn{1}{c}{--} \\
         & \# action & -2.0000 & -3.2188 & -2.9167 & 1.7311 & 13.1152 & 13.5030 & 0.3879 \\
         & \# token  & -275.0000 & 102.6562 & -218.0000 & 51.0756 & 298.4848 & 336.0424 & 37.5576 \\
        \midrule
        \multirow{3}{*}{DeepSeek-V3.1} & \# cases  & 5 & 35 & 14 & 111 & 165 & 165 & \multicolumn{1}{c}{--} \\
         & \# action & 6.4000 & -3.6000 & -3.5000 & 1.5135 & 14.2667 & 14.4182 & 0.1515 \\
         & \# token  & 1270.3333 & -238.9238 & -78.3571 & 303.9790 & 947.3354 & 1132.9960 & 185.6606 \\
        \midrule
        \multirow{3}{*}{Mistral-Large-3} & \# cases  & \loss{\tablenum{5}} & \loss{\tablenum{27}} & \loss{\tablenum{9}} & \loss{\tablenum{124}} & 165 & 165 & \multicolumn{1}{c}{--} \\
         & \# action & 1.4000 & 0.5556 & -0.5556 & -0.8065 & 11.4485 & 10.9455 & -0.5030 \\
         & \# token  & 1115.8000 & 165.1358 & 6.0000 & 36.4113 & 2563.4222 & 2651.9475 & 88.5253 \\
        \bottomrule
    \end{tabular}
    \end{adjustbox}
    \caption{\textbf{Efficiency breakdown under the addition scenario.} We report action and token statistics grouped by paired outcomes between the no-interruption baseline and the interrupted run: S/F denotes baseline success and interruption failure; F/S denotes baseline failure and interruption success; S/S and F/F denote both succeed and both fail, respectively. \textbf{No-int} and \textbf{With-int} are averages over all episodes, and \textbf{Avg.} denotes the overall difference (With-int $-$ No-int). Efficiency breakdowns for the revision and retraction scenarios are provided in Appendix \ref{appendix:efficiency_revision} and Appendix \ref{appendix:efficiency_retraction}.}
    \label{tab:efficiency_addition}
\end{table*}

\subsection{Efficiency of Adaptation}
\label{sec:results:q2}
Success alone does not reveal whether an agent adapts efficiently. We therefore measure both token cost and action cost, and analyze whether models are truly reusing pre-interruption progress or effectively restarting from scratch.

\paragraph{Token and action efficiency.}
For each episode, we record output token usage $T$ and
action count $A$. We compare an interrupted run with update to a matched baseline run with no interruption constructed from the same underlying original task. In Table~\ref{tab:efficiency_addition}, we primarily report absolute differences \(\Delta_T = T_{\text{int}} - T_{\text{base}}\) and \(\Delta_A = A_{\text{int}} - A_{\text{base}}\), so that positive values correspond to higher post-update cost.

\paragraph{Paired-outcome quadrants.}
To disentangle the effect of an update on \emph{outcome} from its effect on \emph{cost}, we assign each episode a paired label \((o_{\text{base}}, o_{\text{int}})\in\{\mathrm{S},\mathrm{F}\}^2\), where \(o_{\text{base}}\) and \(o_{\text{int}}\) denote success (\(\mathrm{S}\)) or
failure (\(\mathrm{F}\)) of the matched baseline run and the interrupted run, respectively. This yields four exhaustive and mutually exclusive outcome quadrants:

\begin{itemize}
    \item S/F (baseline succeeds, interrupted fails): the update is
    \emph{outcome-degrading} for this episode.
    \item F/S (baseline fails, interrupted succeeds): the update is
    \emph{outcome-improving} for this episode.
    \item S/S (both succeed): compare token/action costs under matched
    successful outcomes.
    \item F/F (both fail): analyze whether the update induces fail-fast
    behavior versus prolonged but unproductive interaction.
\end{itemize}
This breakdown reveals whether efficiency is coupled to success and whether models adapt by genuine reuse or implicit restarts.

\noindent\textbf{Token overhead dominates adaptation cost.}
Table~\ref{tab:efficiency_addition} shows that efficiency differences are driven primarily by \emph{token} overhead rather than by additional actions.
Across all models, the overall action delta is small in magnitude, whereas the token delta varies considerably.
For example, under Addition, the average action overhead ranges from $-0.99$ (Opus) to $+0.82$ (Sonnet), while the average token overhead spans from $+37.6$ (Qwen) and $+88.5$ (Mistral) to $+670.9$ (Sonnet) and $+1699.1$ (Haiku).
This suggests that incorporating an additional requirement frequently increases deliberation cost even when the number of actions remains comparable.

\noindent\textbf{F/F failures incur the highest overhead.}
When conditioned on paired outcomes, the largest cost inflation occurs in the both-fail (F/F) quadrant: most models consume more actions and substantially more tokens before ultimately failing, consistent with prolonged but unproductive re-planning.
This pattern is especially pronounced for Haiku ($+2624$ tokens in F/F) and Sonnet ($+1128$ tokens in F/F), and remains evident even for Opus ($+351$ tokens in F/F).

\noindent\textbf{S/F cases are rare but costly.}
In the S/F quadrant (baseline success, interrupted failure), both action and token deltas are predominantly positive across most models (e.g., Sonnet: $+21.75$ actions and $+8584$ tokens; DeepSeek: $+6.40$ actions and $+1270$ tokens), indicating that when an interruption causes a previously successful task to fail, the agent tends to expend more effort before ultimately failing.
Although S/F cases are relatively rare across all models (ranging from 1 to 5 cases), their existence reveals that interruptions can have a detrimental effect: rather than successfully integrating the update, the agent fails to reconcile the new information with its existing plan, resulting in task failure on episodes that would have succeeded without the interruption.

\noindent\textbf{F/S outcomes highlight model capability gaps.}
The F/S quadrant (baseline failure, interrupted success) is the most prevalent outcome category after F/F, and reveals substantial variation across models.
Stronger models show more F/S recoveries (Opus: 70; Sonnet: 55), while open-weight models exhibit fewer recovered cases (DeepSeek: 35; Qwen: 32; Mistral: 27).
Efficiency also differs within recovered episodes: weaker models require extra overhead (e.g., Haiku: $+6.67$ actions and $+1808$ tokens), whereas stronger models can recover with near-zero or negative overhead (Opus: $-2.21$ actions and $-92.7$ tokens).

\noindent\textbf{S/S adaptation is typically lightweight.} Finally, in the both-succeed (S/S) quadrant, both \(\Delta_A\) and \(\Delta_T\) are comparatively small across models, indicating that successful adaptation often reuses pre-interruption progress without requiring a full restart.
Among the strongest models, this manifests as near-zero or negative action overhead, reflecting more direct task completion after integrating the update.

\begin{figure*}[!t]
    \centering
    \includegraphics[width=0.95\textwidth]{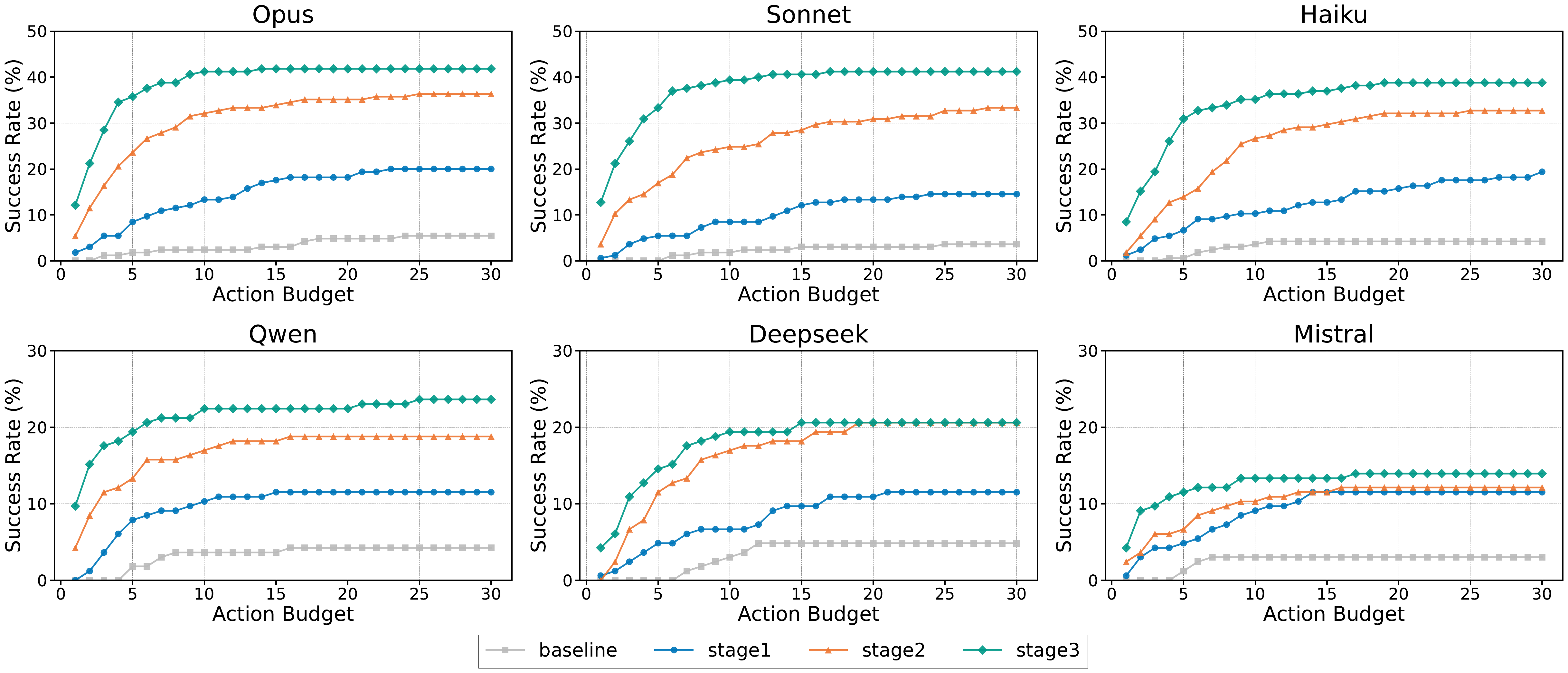}
    \caption{Post-update success curves \(\mathrm{SR}_m(k\mid n)\) for \(n\in\{1,2,3\}\)
    interruptions, aligned at the latest update (so \(k\) counts actions after
    the final interruption). Stage1/2/3 correspond to \(n{=}1/2/3\). The figure
    shows how additional updates shift success upward under a fixed post-update
    budget, with diminishing returns as curves approach a model-specific
    plateau.}
    \label{fig:multi_recovery}
\end{figure*}

% \caption{\textbf{Success under multiple interruption updates.}
% Post-update success curves $\mathrm{SR}_m(k \mid n)$ for $n \in \{1,2,3\}$ interruptions, aligned at the final update so that $k$ counts actions after the last interruption.
% Additional updates shift performance upward under a fixed budget, with diminishing returns as curves approach model-specific plateaus.}

\subsection{Performance Under Multiple Informational Interruptions}
\label{sec:results:q3}
% TODO: adjust to be text wrappted, occupy half textwidth
% \begin{table}[!t]
%     \centering
%     \small
%     \begin{tabular}{lcccc}
%         \toprule
%         Model & 0 int. & 1 int. & 2 int. & 3 int. \\
%         \midrule
%         Claude-Haiku-4.5 & 5.45 & 19.39 & 33.33 & 38.79 \\
%         Claude-Sonnet-4.5 & 4.24 & 15.15 & 33.33 & 41.21 \\
%         Claude-Opus-4.5 & 5.45 & 21.21 & 36.36 & 41.82 \\
%         Qwen3-235B-A22B & 4.24 & 11.52 & 18.79 & 23.64 \\
%         DeepSeek-V3.1 & 4.85 & 12.12 & 21.21 & 20.61 \\
%         Mistral-Large-3 & 3.03 & 11.52 & 12.73 & 13.94 \\
%         \bottomrule
%     \end{tabular}
%     \caption{Converged performance under multiple interruption updates. Success rates under \(n\in\{0,1,2,3\}\) interruptions (more interruptions correspond to more user-provided information).}
%     \label{tab:robustness_multint}
% \end{table}
\begin{wraptable}{r}{0.5\textwidth}
    \centering
    \small
    \setlength{\tabcolsep}{4pt}
    \begin{tabular}{lcccc}
        \toprule
        Model & 0 int. & 1 int. & 2 int. & 3 int. \\
        \midrule
        Claude-Haiku-4.5 & 5.45 & 19.39 & 33.33 & 38.79 \\
        Claude-Sonnet-4.5 & 4.24 & 15.15 & 33.33 & 41.21 \\
        Claude-Opus-4.5 & 5.45 & 21.21 & 36.36 & 41.82 \\
        Qwen3-235B-A22B & 4.24 & 11.52 & 18.79 & 23.64 \\
        DeepSeek-V3.1 & 4.85 & 12.12 & 21.21 & 20.61 \\
        Mistral-Large-3 & 3.03 & 11.52 & 12.73 & 13.94 \\
        \bottomrule
    \end{tabular}
    \caption{Converged performance under multiple interruption updates. Success rates under \(n\in\{0,1,2,3\}\) interruptions (more interruptions correspond to more user-provided information).}
    \label{tab:robustness_multint}
\end{wraptable}
This section evaluates LLM agent performance in episodes with $n \in \{1, 2, 3\}$ interruptions, where each interruption provides an additional user message containing necessary task information.
The agent continues from the same environment state and current trajectory; each update only augments the information available to the model without resetting any prior progress.
We examine how success rate evolves as information is progressively supplied (Figure \ref{fig:multi_recovery}), quantify the marginal gain in success as $n$ increases, assess whether improvements remain stable as updates accumulate, and compare how different LLM backbones leverage multiple interruption updates (Table \ref{tab:robustness_multint}).

% TODO: need to double check this section, need to associate text with Table 2, Figure 3 better
% \emph{stability under accumulation} -> use plain language
% \paragraph{Evaluation Setup.}
% We report the final success rate as a function of $n$ (Table~\ref{tab:robustness_multint}) and characterize multi-interruption performance along two complementary dimensions.
% First, we measure the \emph{marginal information gain}, defined as the change in success rate per additional interruption, which captures how effectively a model utilizes newly provided information.
% Second, we evaluate \emph{stability under accumulation}: whether success increases monotonically with $n$, and whether the relative ranking of models is preserved across different numbers of interruptions.

% more interruption update/information help
\noindent\textbf{Additional user interruption updates generally improve success.}
Across all models, Figure \ref{fig:multi_recovery} and Table~\ref{tab:robustness_multint} shows that providing more information through additional interruptions generally leads to higher success rates.
This supports the premise that incremental clarification or details can help an agent complete long-horizon web tasks without resetting or derailing its ongoing progress.

\noindent\textbf{Claude models improve more consistently and strongly.} Models differ substantially in how predictably and effectively they exploit accumulating information.
The Claude model family exhibits consistent and monotonic improvements across all stages, with Opus achieving the highest overall performance ($41.82\%$ at $n=3$) and Sonnet showing the largest marginal gain between $n=1$ and $n=2$ (an increase of over 18 percentage points).
Notably, all three Claude models roughly double their success rate from $n=1$ to $n=3$, indicating a robust capacity to reliably integrate successive updates and translate them into measurable performance gains.

\noindent\textbf{Open-weight models gain less and are less stable.}
Open-weight models also benefit from additional interruptions, but exhibit smaller marginal gains and less stable improvement patterns.
Among them, Qwen shows gradual but consistent growth, roughly doubling its success rate from $n=1$ to $n=3$, while Mistral improves only marginally across stages, suggesting a limited capacity to exploit progressively revealed task details.
DeepSeek-V3.1 presents a particularly instructive case: it achieves a substantial gain from $n=1$ to $n=2$ but slightly regresses at $n=3$, illustrating a failure mode in which later updates are imperfectly integrated and can interfere with previously incorporated information.

\section{Case Study}
In this section, we present a representative case study from the single-interruption \emph{Addition} setting in the WebArena Map environment. 
% In this setup, the agent is already partway through a long-horizon web task when it receives a mid-trajectory user update that adds necessary information to the original request. Consistent with the benchmark design, the interruption does not reset the agent or the environment; instead, the agent must continue from the current UI state and reconcile its prior progress with the updated intent. 
As illustrated in Figure~\ref{fig:case_study}, the agent begins with the original goal of comparing walking versus driving time to CMU, takes early steps in the web UI such as opening directions, filling the start location, and clicking “Go,” and then receives an interrupt that adds a new constraint: the origin should be Randyland. This update is not a minor wording change; it changes the task’s execution state and therefore the ground-truth comparison. The example is intended to illustrate, at the trajectory level, how successful and unsuccessful interruption handling differ in web interaction.

The contrast is that the failure trace treats the interrupt as a surface-level message: it may acknowledge the update, but it does not fully override the earlier assumption in the environment (the “From” field and computed route remains tied to the old origin), so the final reported comparison is conditioned on stale state. In the correct trace, the agent treats the interrupt as a state-changing event: it explicitly detects intent drift, retracts the previous origin assumption, edits the “From” field to Randyland, triggers a recomputation, and only then reports the updated walking–driving difference. Overall, the case study highlights a key requirement for interruptible agents: effective handling is not just updating the final answer, but ensuring the UI state and intermediate computations are reconciled with the post-interrupt intent before producing any result.

% \section{Discussion}
% \input{sections/6_discussion.tex}

\section{Conclusion}
% We present the first systematic study of interruptible LLM agents in long-horizon, environmentally constrained settings, grounded in realistic web navigation tasks with persistent state changes. Through InterruptBench, which formalizes addition, revision, and retraction as controlled, human-verified interruption scenarios, and a trajectory-grounded simulation and evaluation framework that injects updates at dynamically determined execution points, we enable direct, matched comparisons between interrupted and uninterrupted runs. This setup allows us to jointly measure \emph{effectiveness} and \emph{efficiency}, capturing not only final success but also post-interruption adaptation dynamics and recovery cost. Across six strong backbones, we find that interruption handling remains a persistent bottleneck: agents often continue with stale assumptions, fail to reconcile environment state with updated intent, or produce outputs inconsistent with the revised goal. Although additional user updates can improve success, the gains are uneven and sometimes unstable, especially in multi-turn settings. Overall, our results suggest that robust interruptibility requires stronger mechanisms for state tracking, intent reconciliation, and adaptive execution, highlighting a key challenge for building reliable long-horizon agentic systems.

We present the first systematic study of interruptible LLM agents in long-horizon, environmentally constrained settings, grounded in realistic web navigation tasks with persistent state changes. By formalizing three common interruption types, including addition, revision, and retraction, and instantiating them as controlled, human-verified scenarios in WebArena-Lite, we enable matched comparisons between interrupted and uninterrupted runs. This setting allows us to measure not only final success, but also post-interruption adaptation dynamics and the efficiency cost of recovery. Across six strong backbones, we find that interruption handling remains a persistent bottleneck. Even capable web agents often continue with stale assumptions, fail to reconcile the environment state with the updated intent, or produce answers that are inconsistent with the post-update goal. Although additional user updates generally improve success, the gains are uneven and sometimes unstable. Overall, our results suggest that reliable interruptible agents will require stronger mechanisms for state tracking, intent reconciliation, and error recovery during execution.

% \newpage


\bibliography{colm2026_conference}
\bibliographystyle{colm2026_conference}

%%%%%%%%%%%%%%%%%%%%%%%%%%%%%%%%%%%%%%%%%%%%%%%%%%%%%%%%%%%%%%%%%%%%%%%%%%%%%%%
%%%%%%%%%%%%%%%%%%%%%%%%%%%%%%%%%%%%%%%%%%%%%%%%%%%%%%%%%%%%%%%%%%%%%%%%%%%%%%%
% APPENDIX
%%%%%%%%%%%%%%%%%%%%%%%%%%%%%%%%%%%%%%%%%%%%%%%%%%%%%%%%%%%%%%%%%%%%%%%%%%%%%%%
%%%%%%%%%%%%%%%%%%%%%%%%%%%%%%%%%%%%%%%%%%%%%%%%%%%%%%%%%%%%%%%%%%%%%%%%%%%%%%%
\newpage
\appendix
\onecolumn

% You can have as much text here as you want. The main body must be at most $8$
% pages long. For the final version, one more page can be added. If you want, you
% can use an appendix like this one.

% The $\mathtt{\backslash onecolumn}$ command above can be kept in place if you
% prefer a one-column appendix, or can be removed if you prefer a two-column
% appendix.  Apart from this possible change, the style (font size, spacing,
% margins, page numbering, etc.) should be kept the same as the main body.
%%%%%%%%%%%%%%%%%%%%%%%%%%%%%%%%%%%%%%%%%%%%%%%%%%%%%%%%%%%%%%%%%%%%%%%%%%%%%%%
%%%%%%%%%%%%%%%%%%%%%%%%%%%%%%%%%%%%%%%%%%%%%%%%%%%%%%%%%%%%%%%%%%%%%%%%%%%%%%%

%%%%%%%%%%%%%%%%%%%%%%%%%%%%%%%%%%%%%%%%%%%%%%%%%%%%%%%%%%%%%%%%%%%%%%%%%%%%%%%

% \section{Web Environment \& Dataset Statistics}
% Refer to: (1) Appendix A in WegAgent-R1: https://arxiv.org/pdf/2505.16421 ; (2) RECODE-H, appendix B

% More details about the environment, action space and success checking are provided in Appendix A [refer to WebAgent-R1 Appendix B and WebArena paper]

\renewcommand{\thesection}{\Alph{section}}
\renewcommand{\thefigure}{\thesection.\arabic{figure}}
\renewcommand{\thetable}{\thesection.\arabic{table}}
\setcounter{figure}{0}
\setcounter{table}{0}

\section{Related Work}
% \lipsum[1-3]
%%%%%%%%
% Maybe also take a look at our intro story flow for context info
% % Story Line:
% Background: agent hot -> long-horizon agent is having increasing need and will be widespread in near future (list some recent work on long-horizon agentic benchmark) -> user interruption realistic

% Previous work: only very few work on interruption; one work studies interruption but on purely simple language task: (1) short-horizon; (2)not agentic; (3)state reset/fully immediately reversable; (4) single-turn interruption; 

% This work: agentic long-horizon with environment constraint (web actions & states); we consider three realistic user interrutpion scenarios: (1) addition; (2) revision; (3) retraction. To study this, we constructed our dataset from extremely challenging webarea tasks across 5 diverse domains.

% We benchmarked/analyzed (1) performance/success rate and (2) efficiency (briefly mention the two newly introduced efficiency metrics) + detailed in-depth analysis across sucessfual/failure cases; (3) we also extend our study to multi-turn interrutpion scenarios. We have many interestng findings.

% Contribution: (1) first agentic long horizon interruptable study/benchmark/analysis with env constraints -> webagent/web navigation; (2) extensive experiments & analsysis across 6 strong open/close models across divese interruption scenarios and revealed interesting findings, offering insights for building long-horizon interruptable agents. (maybe split into 3 points)

\subsection{Web-Based Autonomous Agents} % maybe check webagent paper for a better title name
Recent work on web agents shows a consistent evolution across perception, planning, learning, and evaluation. 
To overcome the limitations of raw HTML/DOM representations, prior studies either adopt vision-centric 
grounding to localize actions directly from screen pixels~\citep{lu2024omniparser, gou2025uground, zheng2024seeact}, 
or reduce observation complexity by pruning DOM or accessibility trees to retain task-relevant 
information~\citep{yang2024agentoccam, kerboua2025focusagent}. 
For long-horizon tasks, flat ReAct-style loops have been replaced by more structured reasoning mechanisms, 
including tree search and test-time compute scaling~\citep{dihan2025weboperator, yu2024exact, zhu2025scaling}, 
as well as hierarchical planning and reflection-based control~\citep{li2025hiplan, azam2025reap}. 
Meanwhile, training paradigms have shifted from pure supervised fine-tuning toward environment-feedback-driven 
approaches, such as curriculum reinforcement learning and end-to-end multi-turn 
optimization~\citep{qi2024webrl, wei2025webagent, xue2026evocua}. 
Recent computer-use foundation models demonstrate impressive capabilities in autonomous desktop 
and browser control~\citep{openagi2025lux, fu2025mano}.
Correspondingly, evaluation has moved from static webpages to dynamic, containerized benchmarks, 
established by~\citep{zhou2024webarena} and extended to multimodal understanding in realistic web 
interactions~\citep{he2024webvoyager, koh2024visualwebarena} and end-to-end computer- or 
operating-system-level control~\citep{xie2024osworld}. As agents tackle longer web tasks, user interaction becomes more dynamic, with mid-task updates like additions, revisions, or retractions. Yet most benchmarks assume fixed user intent and uninterrupted execution, leaving agent behavior under realistic interruptions underexplored. Therefore, our paper introduces a benchmark and analysis framework that explicitly evaluates how web agents adapt and recover under such user interruptions to bridge this gap.

\subsection{Human Agent Collaboration System}

Human-agent collaboration systems aim to combine human judgment and domain expertise with the scalability and automation of agents, improving task performance, reliability, and safety in real-world applications~\citep{zou2025survey,zou2025call,shao2025future}. These systems treat humans as integral components that provide additional information~\citep{qian2025userbench}, feedback~\citep{miao2026recode}, or control~\citep{barres2025tau,chen2025embracing} to compensate for the limitations of fully autonomous agents and to improve reliability and safety.
Early work in this area primarily treats humans as reactive helpers, with the agent deciding when and how to request help. \citet{zhangmodeling} train LLMs to predict future conversation turns and to proactively ask clarifying questions when an instruction is ambiguous, thereby improving downstream task performance while keeping humans in a primarily response role. CollabLLM~\citep{wucollabllm} similarly fine-tunes models to move from passive responders to active collaborators that generate clarifying questions and suggestions over multi-turn dialogue, but the user still mainly reacts to model-initiated queries. ReHAC~\citep{feng2024large} further introduces a reinforcement-learning policy that chooses opportune stages for human intervention during complex, multi-step tasks, demonstrating gains from limited, well-timed human input but under fixed task goals and relatively short horizons.
%%%%%%%%%
More recent work shifts toward settings where humans proactively steer, coordinate with, and intervene in agents, rather than being confined to purely reactive assistance. Collaborative Gym~\citep{shao2024collaborative} proposes an open framework where humans and agents engage in bidirectional, non-turn-taking communication while interacting with task environments (e.g., travel planning, related-work writing, tabular analysis), and evaluates both outcome and process metrics. \citet{wu2025large} study how large reasoning models behave when users may interrupt them mid-reasoning or update the problem specification in flight. They reveal severe performance drops and distinct failure modes, such as reasoning leakage, panic, and self-doubt under these user-like interventions.

However, these studies mostly focus on language-only or abstract communication settings and usually treat the task as a fixed, static problem. As a result, they do not capture long-horizon agents acting in persistent environments, where actions update external state and users may change their minds mid-task through additions, revisions, or retractions. In this paper, we treat such user interruptions as a core, realistic form of human–agent collaboration for long-horizon web agents, and introduce a benchmark and analysis framework that jointly evaluates their success, efficiency, and robustness under these environment-bound interruptions.

% \lipsum

% ==================================
% Dataset Statistics
% ==================================

\section{Web Environment \& Dataset Statistics}
\label{appendix:environment_statistics}
% adding

\textbf{WebArena-Lite.} WebArena \citep{zhou2024webarena} is a realistic, self-hostable web environment for developing LLM-based agents. It comprises 812
real-world web tasks spanning diverse domains, including social forum (Reddit), collaborative coding
(GitLab), e-commerce content management system
(CMS), open street map (Map), and online shopping (OneStopShop). WebArena-Lite \citep{koh2024visualwebarena, wei2025webagent} is a curated version of WebArena designed
for more reliable evaluation. It selects 165 representative tasks for human verification as the evaluation set and uses the remaining 647 tasks for training. For each website, the authors summarize the core functionalities and valid items and construct a set of task prototypes and manually implement rule-based solvers using Playwright scripts for each prototype. The corresponding solvers are executed on the websites to collect ground-truth trajectories. For each of the 165 tasks in WebArena-Lite, we synthesized three interruption data corresponding to each interruption scenario (addition, revision, and retraction), using prompts provided in Appendix \ref{appendix:prompt}.

% Refer to: (1) Appendix A in WegAgent-R1: https://arxiv.org/pdf/2505.16421 ; (2) RECODE-H, appendix B

% More details about the environment, action space and success checking are provided in Appendix A [refer to WebAgent-R1 Appendix B and WebArena paper]

\begin{table*}[!t]
    \centering
    \small
    \begin{adjustbox}{width=\textwidth}
    \setlength{\tabcolsep}{4pt}
    \begin{tabular}{ll
        S[table-format=-7.2]
        S[table-format=-7.2]
        S[table-format=-7.2]
        S[table-format=-7.2]
        S[table-format=-7.2]
        S[table-format=-7.2]
        S[table-format=-7.2]
    }
    
        \toprule
        \multicolumn{9}{c}{Scenario: Revision} \\
        \midrule
        Model & Metric & {S/F} & {F/S} & {S/S} & {F/F} & {No-int} & {With-int} & {Avg.} \\
        \midrule

        \multirow{3}{*}{claude-haiku-4-5}
            & \# cases  & 2 & 55 & 6 & 102 & 165 & 165 & \multicolumn{1}{c}{--} \\
            & \# action & 30 & 8.2727 & 4 & 22.0588 & 19.0848 & 16.903 & -2.1818 \\
            & \# token  & 8348.1667 & 2189.4727 & 834.3333 & 7257.0588 & 5551.5313 & 5347.5354 & -203.996 \\
        \midrule

        \multirow{3}{*}{claude-sonnet-4-5}
            & \# cases  & 1 & 62 & 6 & 96 & 165 & 165 & \multicolumn{1}{c}{--} \\
            & \# action & 30 & 8.2097 & 5 & 21.0938 & 18.2848 & 15.7212 & -2.5636 \\
            & \# token  & 12628.6667 & 1975.6129 & 808.5 & 5352.8507 & 4164.6202 & 3962.6747 & -201.9455 \\
        \midrule

        \multirow{3}{*}{claude-opus-4-5}
            & \# cases  & 1 & 86 & 3 & 75 & 165 & 165 & \multicolumn{1}{c}{--} \\
            & \# action & 3 & 7.1977 & 6 & 17.8 & 16.3394 & 11.9697 & -4.3697 \\
            & \# token  & 639 & 1366.1783 & 988.6667 & 2732.6178 & 2761.4303 & 1976.0162 & -785.4141 \\
        \midrule

        \multirow{3}{*}{qwen3-235b-a22b}
            & \# cases  & 0 & 42 & 1 & 122 & 165 & 165 & \multicolumn{1}{c}{--} \\
            & \# action & 0 & 5.6905 & 4 & 21.3852 & 16.5394 & 17.2848 & 0.7455 \\
            & \# token  & 0 & 271.8651 & 73 & 550.1858 & 575.6848 & 476.4485 & -99.2364 \\
        \midrule

        \multirow{3}{*}{deepseek.v3.1}
            & \# cases  & 1 & 41 & 2 & 121 & 165 & 165 & \multicolumn{1}{c}{--} \\
            & \# action & 30 & 8.561 & 3 & 17.2066 & 16 & 14.9636 & -1.0364 \\
            & \# token  & 1455.3333 & 786.3415 & 47 & 1195.4215 & 1130.2182 & 1081.4263 & -48.7919 \\
        \midrule

        \multirow{3}{*}{mistral-large-3}
            & \# cases  & 0 & 25 & 4 & 136 & 165 & 165 & \multicolumn{1}{c}{--} \\
            & \# action & 0 & 6.64 & 11 & 14.7059 & 12.2242 & 13.3939 & 1.1697 \\
            & \# token  & 0 & 3690.7067 & 1337.0833 & 3717.7353 & 2509.7091 & 3655.9273 & 1146.2182 \\
        \bottomrule
    \end{tabular}
    \end{adjustbox}
    \caption{\textbf{Efficiency breakdown under the revision scenario shown above.} We report action and token statistics grouped by paired outcomes between the interrupted run and the no-interruption baseline: S/F denotes interruption success and baseline failure; F/S denotes interruption failure and baseline success; S/S and F/F denote both succeed and both fail, respectively. \textbf{No-int} and \textbf{With-int} are averages over all episodes, and \textbf{Avg.} denotes the difference (With-int $-$ No-int).}
    \label{tab:efficiency_fillme_2}
\end{table*}

% ==================================
% Additional Result I
% ==================================

\section{Efficiency Analysis under Revision Interruptions}
\label{appendix:efficiency_revision}
Table~\ref{tab:efficiency_fillme_2} summarizes paired outcomes and efficiency statistics under the scenario shown above. Across models, outcome disagreements are predominantly concentrated in F/S (interruption failure, baseline success) rather than S/F (interruption success, baseline failure); for instance, Claude-Opus exhibits F/S$=86$ versus S/F$=1$, and Claude-Sonnet shows F/S$=62$ versus S/F$=1$. With respect to efficiency, interruptions reduce the average number of actions for several models (negative \textbf{Avg.} for number action on the Claude family and DeepSeek), and are often accompanied by token reductions (negative \textbf{Avg.} for number of token). This trend is not universal: Qwen shows a slight increase in actions despite fewer tokens, while Mistral incurs substantial additional token usage. These results indicate that the efficiency impact of interruption handling is highly model- and scenario-dependent, and can entail trade-offs between interaction length and generation cost.

% ==================================
% Additional Result II
% ==================================

\section{Efficiency Analysis under Retraction Interruptions}
\label{appendix:efficiency_retraction}
Relative to Table~\ref{tab:efficiency_fillme_2}, Table~\ref{tab:efficiency_fillme_3} places more mass on S/S and exhibits non-trivial S/F counts, suggesting that interruptions can correct a subset of baseline failures in this scenario (while still introducing some F/S cases). Notably, all models exhibit lower average action counts with interruptions (negative \textbf{Avg.} for number of action, ranging from $-1.72$ to $-8.13$), consistent with shorter action sequences when user updates are incorporated. Token usage decreases for most models (e.g., Claude-Opus: \textbf{Avg.} $\approx -1061$ tokens; DeepSeek: $\approx -330$ tokens), although Haiku shows a marginal increase (\textbf{Avg.} $=+8.8$). Overall, these results suggest that informative interruptions can reduce action steps and often reduce token consumption, but the magnitude and even direction of token changes remain sensitive to both the model and the scenario.

\begin{table*}[!t]
    \centering
    \small
    \begin{adjustbox}{width=\textwidth}
    \setlength{\tabcolsep}{4pt}
    \begin{tabular}{ll
        S[table-format=-7.2]
        S[table-format=-7.2]
        S[table-format=-7.2]
        S[table-format=-7.2]
        S[table-format=-7.2]
        S[table-format=-7.2]
        S[table-format=-7.2]
    }
        \toprule
        \multicolumn{9}{c}{Scenario: Retraction} \\
        \midrule
        Model & Metric & {S/F} & {F/S} & {S/S} & {F/F} & {No-int} & {With-int} & {Avg.} \\
        \midrule

        \multirow{3}{*}{claude-haiku-4-5}
            & \# cases  & 5 & 38 & 24 & 98 & 165 & 165 & \multicolumn{1}{c}{--} \\
            & \# action & 14.4 & 8.5263 & 7.625 & 23.8265 & 24.3576 & 17.6606 & -6.697 \\
            & \# token  & 7862.8667 & 2976.5526 & 3812.5972 & 9956.602 & 7383.1556 & 7391.9556 & 8.8 \\
        \midrule

        \multirow{3}{*}{claude-sonnet-4-5}
            & \# cases  & 8 & 29 & 35 & 93 & 165 & 165 & \multicolumn{1}{c}{--} \\
            & \# action & 19.375 & 8.7241 & 6.5143 & 24.1613 & 22.8848 & 17.4727 & -5.4121 \\
            & \# token  & 5444.875 & 2454.5287 & 1793.0762 & 8384.9534 & 5938.8343 & 5801.8101 & -137.0242 \\
        \midrule

        \multirow{3}{*}{claude-opus-4-5}
            & \# cases  & 6 & 41 & 44 & 74 & 165 & 165 & \multicolumn{1}{c}{--} \\
            & \# action & 6.8333 & 6.9756 & 7.0682 & 18.3378 & 20.2182 & 12.0909 & -8.1273 \\
            & \# token  & 1098.7778 & 1822.9512 & 1755.697 & 3884.5721 & 3764.0263 & 2703.2889 & -1060.7374 \\
        \midrule

        \multirow{3}{*}{qwen3-235b-a22b}
            & \# cases  & 9 & 23 & 24 & 109 & 165 & 165 & \multicolumn{1}{c}{--} \\
            & \# action & 15.3333 & 5.4348 & 3.9167 & 19.7248 & 21.0303 & 15.1939 & -5.8364 \\
            & \# token  & 343 & 137.7826 & 101.625 & 604.4526 & 607.5697 & 452.002 & -155.5677 \\
        \midrule

        \multirow{3}{*}{deepseek.v3.1}
            & \# cases  & 7 & 29 & 20 & 109 & 165 & 165 & \multicolumn{1}{c}{--} \\
            & \# action & 16.4286 & 9.1034 & 5.9 & 15.0826 & 19.4242 & 12.9758 & -6.4485 \\
            & \# token  & 938.619 & 918.1379 & 548.75 & 1253.2294 & 1425.7717 & 1095.596 & -330.1758 \\
        \midrule

        \multirow{3}{*}{mistral-large-3}
            & \# cases  & 10 & 21 & 11 & 123 & 165 & 165 & \multicolumn{1}{c}{--} \\
            & \# action & 9.6 & 8.381 & 15.0909 & 13.5447 & 14.4727 & 12.7515 & -1.7212 \\
            & \# token  & 2249.0333 & 2386.4286 & 3136.4242 & 3459.3496 & 3514.4929 & 3227.9152 & -286.5778 \\
        \bottomrule
    \end{tabular}
    \end{adjustbox}
    \caption{\textbf{Efficiency breakdown under the retraction scenario shown above.} We report action and token statistics grouped by paired outcomes between the interrupted run and the no-interruption baseline: S/F denotes interruption success and baseline failure; F/S denotes interruption failure and baseline success; S/S and F/F denote both succeed and both fail, respectively. \textbf{No-int} and \textbf{With-int} are averages over all episodes, and \textbf{Avg.} denotes the difference (With-int $-$ No-int).}
    \label{tab:efficiency_fillme_3}
\end{table*}

% ==================================
% Evaluations on Different Interruption Position
% ==================================

\section{Evaluation on Different Interruption Position}
\label{appendix:eval_position}
In Table \ref{tab:interrupt_position}, we evaluate model sensitivity to interruption position by injecting updates at $\{0.2, 0.4, 0.6, 0.8\}$ of the baseline trajectory. This probes how the amount of committed intermediate state prior to interruption affects post-update adaptation. 
Overall, we observe a scale-dependent pattern. Large-scale models (e.g., Claude-Opus-4.5) achieve their best overall success when interruptions occur at later positions, indicating effective reuse of pre-interruption progress and localized trajectory repair. In contrast, smaller and open-weight models (e.g., Claude-Haiku, DeepSeek-V3.1) peak under early interruption and degrade as interruption is delayed, suggesting higher sensitivity to stale assumptions and weaker reconciliation of environment state with updated intent. 
Late interruptions require not only update integration but also correction of persistent UI state and previously executed actions. Larger models appear capable of repairing committed trajectories, whereas relatively smaller models exhibit trajectory entanglement under delayed updates.

\begin{table*}[!t]
\centering
\small
\begin{adjustbox}{width=\textwidth}
\setlength{\tabcolsep}{15pt}
\begin{tabular}{l c ccccc c}
\toprule
Model & Interrupt Position & Reddit & GitLab & CMS & Map & Shopping & Overall \\
\midrule
\multirow{5}{*}{Claude-Haiku-4.5}
& baseline & 5.30 & 20.00 & 11.40 & 0.00 & 20.00 & 12.12 \\
& 0.2 & 31.60 & 50.00 & 54.30 & 23.10 & 51.10 & \textbf{43.64} \\
& 0.4 & 26.30 & 36.70 & 37.10 & 15.40 & 33.30 & 29.70 \\
& 0.6 & 15.80 & 33.30 & 25.70 & 23.10 & 35.60 & 27.88 \\
& 0.8 & 21.10 & 36.70 & 17.10 & 15.40 & 31.10 & 24.24 \\
\midrule
\multirow{5}{*}{Claude-Sonnet-4.5}
& baseline & 0.00 & 20.00 & 2.90 & 7.70 & 28.90 & 13.33 \\
& 0.2 & 47.40 & 66.70 & 60.00 & 19.20 & 46.70 & \textbf{47.27} \\
& 0.4 & 47.40 & 53.30 & 57.10 & 26.90 & 42.20 & 44.24 \\
& 0.6 & 47.40 & 53.30 & 57.10 & 19.20 & 35.60 & 41.21 \\
& 0.8 & 52.60 & 53.30 & 48.60 & 23.10 & 42.20 & 42.42 \\
\midrule
\multirow{5}{*}{Qwen3-235B-A22B}
& baseline & 5.30 & 16.70 & 2.90 & 0.00 & 11.10 & 7.27 \\
& 0.2 & 26.30 & 43.30 & 45.70 & 11.50 & 28.90 & 31.52 \\
& 0.4 & 15.80 & 46.70 & 51.40 & 7.70 & 33.30 & \textbf{32.73} \\
& 0.6 & 15.80 & 36.70 & 40.00 & 7.70 & 33.30 & 27.27 \\
& 0.8 & 5.30 & 33.30 & 48.60 & 11.50 & 33.30 & 27.88 \\
\midrule
\multirow{5}{*}{DeepSeek-V3.1}
& baseline & 0.00 & 23.30 & 8.60 & 0.00 & 11.10 & 9.09 \\
& 0.2 & 26.30 & 33.30 & 40.00 & 15.40 & 31.10 & 30.30 \\
& 0.4 & 26.30 & 36.70 & 37.10 & 15.40 & 33.30 & \textbf{29.70} \\
& 0.6 & 15.80 & 33.30 & 25.70 & 23.10 & 35.60 & 27.88 \\
& 0.8 & 21.10 & 36.70 & 17.10 & 15.40 & 31.10 & 24.24 \\
\midrule
\multirow{5}{*}{Mistral-Large-3}
& baseline & 0.00 & 13.30 & 8.60 & 0.00 & 11.10 & 7.27 \\
& 0.2 & 10.50 & 26.70 & 25.70 & 11.50 & 33.30 & 22.42 \\
& 0.4 & 21.10 & 23.30 & 20.00 & 11.50 & 24.40 & 19.39 \\
& 0.6 & 10.50 & 30.00 & 20.00 & 11.50 & 31.10 & 21.21 \\
& 0.8 & 5.30 & 33.30 & 48.60 & 11.50 & 33.30 & \textbf{27.88} \\
\midrule
\multirow{5}{*}{Claude-Opus-4.5}
& baseline & 10.50 & 23.30 & 8.60 & 11.50 & 17.80 & 13.94 \\
& 0.2 & 63.20 & 50.00 & 60.00 & 23.10 & 48.90 & 47.27 \\
& 0.4 & 63.20 & 60.00 & 68.60 & 34.60 & 60.00 & 55.76 \\
& 0.6 & 52.60 & 66.70 & 71.40 & 30.80 & 55.60 & 54.55 \\
& 0.8 & 57.90 & 63.30 & 68.60 & 38.50 & 62.20 & \textbf{56.97} \\
\bottomrule
\end{tabular}
\end{adjustbox}
\caption{\textbf{Task success rate under different interruption positions.} ``baseline'' denotes no-interruption runs that do \emph{not} receive the update. Overall, large-scale models improve under later interruptions, while relatively smaller models degrade as interruption is delayed, indicating higher sensitivity to stale/noisy intermediate state.}
\label{tab:interrupt_position}
\end{table*}

% ==================================
% Prompts
% ==================================

\section{Prompt for Data Synthesis}
\label{appendix:prompt}
To construct user query for evaluating interruptible agents, we designed four distinct data synthesis strategies that simulate realistic user behavior patterns during web navigation tasks. Each strategy targets a specific type of user interruption: adding missing details, revising incorrect information, retracting unnecessary constraints, or combining multiple update types in a multi-turn interaction. We employed Claude-Opus-4.5 to generate these synthetic variations from WebArena, ensuring that the transformed queries keep semantic equivalence with original intent while reflecting natural user communication patterns.

% --- Minimalist Grayscale Professional ---
\definecolor{ProGray}{HTML}{4a5568}
\definecolor{ProLight}{HTML}{f7fafc}
\definecolor{ProBorder}{HTML}{e2e8f0}
\definecolor{ProAccent}{HTML}{718096}
\definecolor{ProDark}{HTML}{2d3748}

\lstdefinestyle{proprompt}{
  basicstyle=\ttfamily\footnotesize\color{ProDark},
  columns=fullflexible,
  breaklines=true,
  breakindent=0pt,
  showstringspaces=false,
  keepspaces=true,
  upquote=true,
  xleftmargin=0pt,
  resetmargins=true,
  backgroundcolor=\color{ProLight},
  commentstyle=\color{ProAccent!70}\itshape,
  keywordstyle=\color{ProGray}\bfseries,
  stringstyle=\color{ProAccent}
}

\newtcblisting{PromptBox}[1]{%
  enhanced,
  breakable,
  listing only,
  listing options={style=proprompt},
  colback=white,
  colframe=ProBorder,
  boxrule=1pt,
  arc=2pt,
  left=6mm, right=6mm, top=2mm, bottom=4mm,
  before skip=12pt,
  after skip=12pt,
  borderline west={4pt}{0pt}{ProGray},
  title={\sffamily\bfseries #1},
  fonttitle=\normalsize,
  coltitle=white,
  colbacktitle=ProGray,
  boxed title style={
    boxrule=0pt,
    sharp corners,
    left=3mm, right=3mm, top=1.5mm, bottom=1.5mm,
  },
  attach boxed title to top left={xshift=6mm, yshift=-2.5mm},
  drop fuzzy shadow
}

\subsection{Addition Strategy}
\label{appendix:prompt_addition}
This prompt generates scenarios where users incrementally provide missing details to their initial incomplete query. The model transforms a complete task specification into a less detailed initial query and produces natural follow-up updates that add the missing information. This simulates common real-world scenarios where users start with a vague request and progressively clarify their requirements.

\begin{PromptBox}{Data Synthesis--Addition}
Scenario: We are simulating an interruption scenario where a user may provide updates to its initial query for an agent.    

Task: Given the task_metadata, now revise the intent query by making {num_update} updates. 

Requirement: 
1. The transformed_initial_intent combined with the updates should have the same meaning as the original intent query and result in the same ground truth answers. 
2. Express the updates in a realistic user/customer tone and be concise.
3. Each update should be essential-lacking any one update should affect the ground truth answers.


#####
Output: 
[{
task_id: {task_id}
intent: {original_intent}
transformed_initial_intent: {transformed_initial_intent}
updates: [{update_1}, {update_2}, {update_3}]
}]


######
Input: 
num_update: {num_of_update}
task_metadata: {task_metadata}
\end{PromptBox}

\subsection{Revision Strategy}
\label{appendix:revision}
This prompt creates scenarios where users initially provide incorrect information and subsequently correct their mistakes. The model generates plausible but incorrect details in the initial query (such as wrong dates, products, or quantities) and produces natural correction updates that fix these errors to recover the original intent. This reflects real-world situations where users realize and rectify their initial mistakes during interaction.

\begin{PromptBox}{Data Synthesis--Revision}
Scenario: We are simulating a modification interruption scenario where a user initially provides an incorrect or mistaken query and then follows up with corrections/modifications to fix the errors.

Task: Given the task_metadata, revise the intent query by:
Creating a transformed_initial_intent that changes {num_update} critical detail(s) from the original intent (e.g., wrong date, wrong product, wrong quantity, wrong location, etc.)
Generating corresponding updates that correct those mistaken details to recover the original intent.

Requirements:
1. The transformed_initial_intent should contain plausible but incorrect details compared to the original intent.
2. The transformed_initial_intent combined with the updates should have the same meaning as the original intent and result in the same ground truth answers.
3. Express the updates as realistic corrections in a natural user/customer tone.
4. Each correction should be essential-lacking any one update should result in different ground truth answers than the original intent.


#####
Output: 
[{task_i: {task_id}
intent: {original_intent}
transformed_initial_intent: {transformed_initial_intent}
updates: [{update_1}, {update_2}, {update_3}]
}]


######
Input: 
num_update: {num_of_update}
task_metadata: {task_metadata}
\end{PromptBox}

\subsection{Retraction Strategy}
\label{appendix:retraction}
This prompt generates scenarios where users initially over-specify their requirements with extra constraints or details, then remove these unnecessary elements. The model adds additional filters, requirements, or specifications to the original query and produces natural retraction updates that remove these extras. This simulates situations where users start with overly restrictive criteria and progressively relax their constraints to broaden their search or simplify their request.

\begin{PromptBox}{Data Synthesis--Retraction}
Scenario: We are simulating a retraction interruption scenario where a user initially provides a query with extra details/constraints and then follows up to remove or retract some of those details.

Task: Given the task_metadata, revise the intent query by:
Creating a transformed_initial_intent that adds {num_update} additional critical detail(s)/constraint(s) to the original intent (e.g., extra filters, additional requirements, more specifications, etc.)
Generating corresponding updates that remove/retract those extra details to recover the original intent.

Requirements:
1. The transformed_initial_intent should contain additional plausible details/constraints beyond the original intent.
2. The transformed_initial_intent combined with the updates should have the same meaning as the original intent and result in the same ground truth answers.
3. Express the updates as realistic retractions in a natural user/customer tone.
4. Each retraction should be essential-lacking any one update should result in different ground truth answers than the original intent.


#####
Output: 
[{
task_id: {task_id}
intent: {original_intent}
transformed_initial_intent: {transformed_initial_intent}
updates: [{update_1}, {update_2}, {update_3}]
}]


######
Input: 
num_update: {num_of_update}
task_metadata: {task_metadata}
\end{PromptBox}

\subsection{Multi-Turn Mixed Strategy}
\label{appendix_multiturn}
This prompt creates the most complex and realistic scenarios by combining all three update types (addition, revision, and retraction) within a single interaction sequence. The model generates an initial query that is simultaneously incomplete, partially incorrect, and over-constrained, then produces a series of mixed updates to recover the original intent. The updates are randomized in order and explicitly labeled by type, simulating natural multi-turn conversations where users iteratively refine their requests through various types of modifications.

\begin{PromptBox}{Data Synthesis--Multi Turn}
Scenario: We are simulating a mixed multi-turn interruption scenario where a user initially provides a query and then follows up with a combination of:
Additions: Providing missing details that were not in the initial query
Modifications: Correcting mistaken/incorrect details from the initial query
Retractions: Removing extra details/constraints that were initially included


Task: Given the task_metadata, revise the intent query by:
Creating a transformed_initial_intent that differs from the original intent through a combination of:
Missing critical detail(s) (to be added later)
Incorrect detail(s) (to be corrected later)
Extra detail(s)/constraint(s) (to be retracted later)
Generating {num_update} corresponding updates that recover the original intent through additions, modifications, and/or retractions.

Requirements:
1. The transformed_initial_intent should be a plausible but incomplete/incorrect/over-specified/mixed version of the original intent.
2. The transformed_initial_intent combined with all updates should have the same meaning as the original intent and result in the same ground truth answers.
3. Express the updates in a realistic user/customer tone and be concise.
Each update should be essential-lacking any one update should affect the ground truth answers.
4. The updates should include a mix of at three different types (addition, modification, retraction). The order of types should be randomized.
Clearly label each update with its type in the updates_type array.


#####
Output: 
[{
task_id: {task_id}
intent: {original_intent}
transformed_initial_intent: {transformed_initial_intent}
"updates_type": ["{type_1}", "{type_2}", "{type_3}"],
updates: [{update_1}, {update_2}, {update_3}]
}]


######
Input: 
num_update: {num_of_update}
task_metadata: {task_metadata}

\end{PromptBox}

\newpage

\begin{figure*}[t]
    \centering
    %\hspace*{2mm}
    \includegraphics[width=\textwidth]{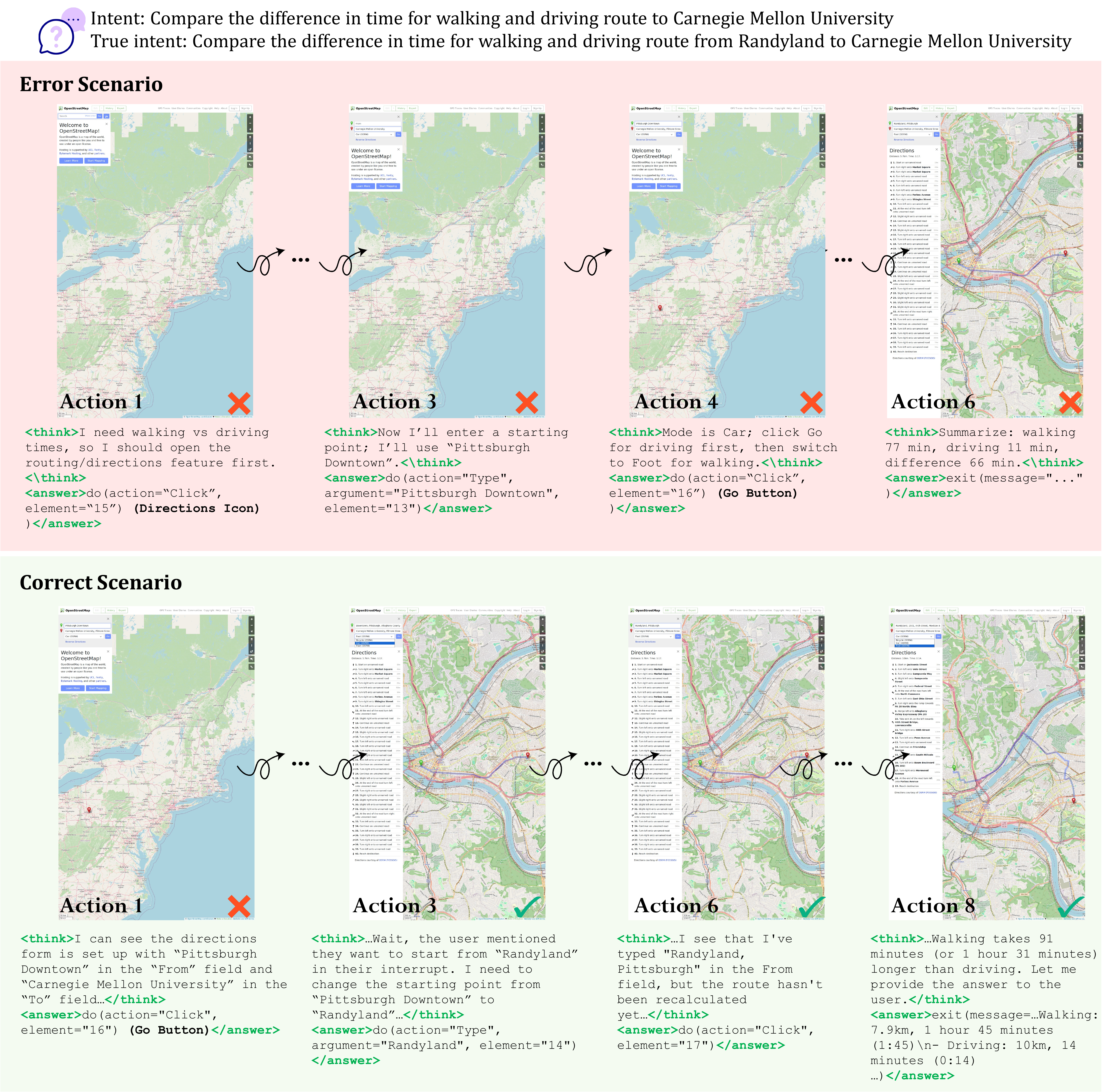} % example-image-duck
    \caption{User interruption with additional intent in WebArena. The agent initially plans a walking vs. driving comparison from Pittsburgh Downtown → Carnegie Mellon University, but a mid-trajectory user interrupt adds a new constraint (start from Randyland); we contrast an error trace that fails to fully retract/update the plan with a correct trace that revises the origin, recomputes routes, and returns the updated time difference.}
    \label{fig:case_study}
\end{figure*}

\end{document}